%% file: acl_latex.tex
\pdfoutput=1

\documentclass[11pt]{article}

\usepackage[final]{acl}
\usepackage{amsmath}
\usepackage{times}
\usepackage{latexsym}
\usepackage{microtype}
\usepackage{hyperref}
\usepackage{url}
\usepackage{booktabs}
\usepackage{graphicx} 
\usepackage{lineno}
\usepackage{subcaption}
\usepackage{caption}
\usepackage{multirow}
\usepackage{wrapfig}
\usepackage{enumitem}
\usepackage{color,amsfonts}
\usepackage{array}
\usepackage{algorithm}
\usepackage{algorithmic}
\usepackage{cleveref}
\usepackage{caption}
\usepackage{makecell}
\usepackage[most]{tcolorbox}
\usepackage{alltt}
\usepackage{courier} 
\usepackage{fvextra} 
\usepackage[utf8]{inputenc}
\usepackage[T1]{fontenc}
\usepackage[x11names]{xcolor}
\definecolor{darkblue}{rgb}{0, 0, 0.5}
\hypersetup{colorlinks=true, citecolor=darkblue, linkcolor=darkblue, urlcolor=darkblue}
\input{ACL_content/notation}
\title{\textsc{ConGrad}: Conflicting Gradient Filtering for Multilingual Preference Alignment}

\author{
  Jiangnan Li\textsuperscript{1}, Thuy-Trang Vu\textsuperscript{1}, Christian Herold\textsuperscript{2}, Amirhossein Tebbifakhr\textsuperscript{2},\\ \textbf{Shahram Khadivi\textsuperscript{2}, Gholamreza Haffari\textsuperscript{1}} \\
  \textsuperscript{1}Department of Data Science and AI, Monash University, Australia \\
  \textsuperscript{2}eBay Inc. \\
  \texttt{\{first.last, trang.vu1\}@monash.edu,}\\ 
  \texttt{\{cherold, atebbifakhr, skhadivi\}@ebay.com}
}

\begin{document}
\maketitle
\begin{abstract}
\input{ACL_content/abstract}

\end{abstract}

\input{ACL_content/1-intro}

\input{ACL_content/2-background}
\input{ACL_content/3-method}
\input{ACL_content/experiment}

\input{ACL_content/5-analysis}

\input{ACL_content/7-conclusion}

\input{ACL_content/limitation}

\bibliography{custom}

\appendix

\section{Example Appendix}
\label{sec:appendix}
\input{ACL_content/Appendix}

\end{document}

%% file: ACL_content/notation.tex
\usepackage{amsthm,amsmath,amsfonts,xspace}

\def\eg{{\em e.g.,}\xspace}

\newcommand\cD{\mathcal{D}}

\newcommand\cL{\mathcal{L}}
\newcommand\cM{\mathcal{M}}

\newcommand\cR{\mathcal{R}}

\newcommand\cX{\mathcal{X}}
\newcommand\cY{\mathcal{Y}}

\newcommand\E{\mathbb{E}}

\newcommand\ychosen{y_c}
\newcommand\yrejected{y_r}
\newcommand\yichosen{y_{i,c}}
\newcommand\yirejected{y_{i,r}}
\newcommand\grad{\mathbf{g}}
\newcommand\gradpc{\mathbf{g}_{\textrm{pc}}}

%% file: ACL_content/abstract.tex
Naive joint training of large language models (LLMs) for multilingual preference alignment can suffer from \emph{negative interference}. This is a known issue in \emph{multilingual training}, where conflicting objectives degrade overall performance. 
However, the impact of this phenomenon in the context of \emph{multilingual preference alignment} remains largely underexplored.
To address this issue, we propose \textsc{ConGrad}, an effective and scalable filtering method that mitigates this interference by identifying and selecting preference samples that exhibit high cross-lingual affinity.
Based on principles of multi-objective optimization, our approach computes an aggregated, cross-lingually beneficial gradient direction and uses this to filter for samples whose individual gradients align with this consensus direction.
To ensure scalability for LLMs, we incorporate a \emph{sublinear gradient compression} strategy that reduces memory overhead during gradient accumulation.
We integrate \textsc{ConGrad} into a self-rewarding framework and evaluate on \texttt{LLaMA3-8B} and \texttt{Gemma2-2B} across 10 languages. 
Results show that \textsc{ConGrad} consistently outperforms strong baselines in both seen and unseen languages, with minimal alignment tax.
\footnote{Code will be released upon acceptance}

%% file: ACL_content/1-intro.tex
\section{Introduction}
Preference alignment has emerged as a pivotal post-training technique for aligning large language models (LLMs) with human values and intentions \citep{ouyang2022training, touvron2023llama}. It has driven significant performance improvements across multiple natural language processing (NLP) tasks such as summarisation \citep{ziegler2019fine}, reasoning \citep{pang2024iterative, spin2024}, and instruction following \citep{Bai2022Traininghelpfulharmless, ouyang2022training}. 
However, the progress in preference alignment research remains predominantly English-centric due to the scarcity of high-quality human preference data for non-English languages and the prohibitive costs of human annotation. Previous research observes that state-of-the-art aligned LLMs often overfit to English capabilities with noticeable degraded alignment and odd behaviours in less-represented languages \citep{schwartz2022towards,10.1145/3582269.3615599, khondaker-etal-2023-gptaraeval, vashishtha-etal-2023-evaluating, deng2023multilingual}.

To circumvent the need for costly human annotation in less-represented languages, recent efforts have turned to synthetic data generation \citep{hurst2024gpt}, often using powerful but proprietary models such as  \texttt{OpenAI GPT-4o} \citep{hurst2024gpt} as judges to score responses \citep{dubois2023alpacafarm,rlaif,ultrafeedback} or employing self-rewarding methods where a model iteratively generates and evaluates its own outputs \citep{Yuan2024Selfrewardinglanguage}. While promising, these strategies frequently rely on translating English-centric datasets \citep{Lai2023OkapiInstructiontuned, chen-etal-2024-monolingual}, a process that can introduce subtle artifacts and fails to generate diverse samples, which are important to robust model performances \citep{kirk2024understanding}. Once constructed, these multilingual preference datasets are typically used to jointly train aligned multilingual LLMs \citep{Dang2024RLHFCanSpeak}.

Nevertheless, it is well-known in multilingual machine translation and pre-training that naive optimization in multilingual tasks often exhibits \textit{negative interference}, where conflicting per-language objectives often lead to sub-optimal performance \citep{Wang2020NegativeInterferenceMultilingual}. Similar to multilingual machine translation, multilingual preference alignment is also fundamentally a multi-objective learning problem and prone to the negative interference issue \citep{Wang2020NegativeInterferenceMultilingual, wang2021gradient}. While this issue has been widely studied in multilingual machine translation and pretraining \citep{arivazhagan2019massively,Wang2020NegativeInterferenceMultilingual, conneau-etal-2020-unsupervised, Wu2021UncertaintyAwareBalancing,choi2023order,wu-etal-2024-mixture-skills}, the impact of negative interference in multilingual preference alignment remains largely unexplored.

This paper directly addresses this gap. We propose \textsc{ConGrad} (CONflicting GRADient filtering), an effective and scalable method to mitigate negative interference in multilingual preference alignment. Our approach is motivated by the insight that gradient conflicts are strong indicators of task interference~\citep{sener2018multi, Yu2020Gradientsurgerymulti}. \textsc{ConGrad} filters the training data by identifying and retaining only those preference samples whose gradients align with a consensus direction beneficial to all languages to alleviate negative interference. This is achieved by first aggregating accumulated exponential moving average (EMA) gradients across languages and resolving their conflicts, and then selecting the top-$k$ samples with the highest similarity to this de-conflicted, consensus gradient. Unlike traditional algorithms that directly modify gradients~\citep{sener2018multi, Yu2020Gradientsurgerymulti, wang2021gradient}, our method filters data samples to influence optimization indirectly. This approach is better suited for LLMs as it avoids the significant memory/computational overhead and potential training instability associated with direct gradient manipulation~\citep{kurin2022defense}. Besides, to ensure this method is more scalable for contemporary LLMs, we incorporate a sublinear gradient compression strategy based on subspace iteration~\citep{bathe1972large} that makes storing and processing gradients memory-efficient.

We integrate our proposed \textsc{ConGrad} filtering method into the self-rewarding framework \citep{Yuan2024Selfrewardinglanguage} with Direct Preference Optimization (DPO) training \citep{Rafailov2023Directpreferenceoptimization}. To evaluate its effectiveness, we conduct experiments on two LLMs of different scales: \texttt{Llama3-8B} \citep{grattafiori2024llama} and \texttt{Gemma2-2B} \citep{team2024gemma} across 10 languages from different language families and writing scripts. The experimental results demonstrate the importance of high-quality preference data, where most filtering methods outperform the baselines without data filtering. Our proposed \textsc{ConGrad} method consistently outperforms strong filtering baselines on the multilingual instruction following benchmark for both languages seen during training, generalised well to unseen languages and low-resource languages, without significant alignment tax \citep{lin-etal-2024-mitigating}.

In summary, our contributions are as follows.
\begin{itemize}[noitemsep,topsep=0pt,parsep=0pt,partopsep=0pt,leftmargin=13pt]
\item We propose a conflict-aware self-rewarding algorithm for multilingual preference alignment, which constructs preference data without relying on external annotations and enables iterative self-improvement across languages.

\item We introduce a gradient-level filtering strategy based on PCGrad that selects preference samples with high cross-lingual affinity, effectively mitigating negative interference during multilingual training.

\item We conduct comprehensive experiments on \texttt{Llama3-8B} and \texttt{Gemma2-2B}, showing our method significantly improves underrepresented language while preserving dominant language capabilities and generalizes to unseen languages with minimal alignment tax. 
\end{itemize}

%% file: ACL_content/2-background.tex
\section{Preliminaries}

\paragraph{Multilingual Preference Alignment}
In this paper, we study the problem of multilingual preference alignment. For a given set of languages $L$, each language $l \in L$ is associated with a preference dataset $\cD^{l} = \{(x^{l}_i, \yichosen^l,\yirejected^l)\}_{i=1}^{N^l}$ where $N^l$ is the dataset size. Each data point is a tuple of prompt input $x^{l}_i$, chosen response $\yichosen^l$ and rejected response $\yirejected^l$. For brevity, we omit the index $i$ when it is clear from the context.

The goal of mono-lingual alignment is to align the LLM $\cM_\theta$, parameterised by $\theta$, using preference optimisation methods such as reinforcement learning with human feedback (RLHF), \eg PPO~\citep{schulman2017proximal} or Direct Preference Optimisation (DPO) \citep{Rafailov2023Directpreferenceoptimization}. We adopt DPO due to its simplicity and strong empirical performance. DPO directly minimises the negative log probability of preferring the chosen response $\ychosen$ over the rejected response $\yrejected$ given prompt $x$,
\begin{equation}
\label{eq-dpo-perfect}
\begin{aligned}
  & \cL_{\text{DPO}}(\cM_\theta, \cD) ={} \\
  & -\E_{(x, \ychosen, \yrejected) \sim \cD} \Bigg[ \log \sigma \bigg( \beta \log\frac{\cM_{\theta}(\ychosen|x)}{\cM_{\text{ref}}(\ychosen|x)}- \\
  & \hphantom{-\E_{(x, \ychosen, \yrejected) \sim \cD} \Bigg[ \log \sigma \bigg(}  \beta \log\frac{\cM_{\theta}(\yrejected|x)}{\cM_{\text{ref}}(\yrejected|x)} \bigg) \Bigg]
\end{aligned}
\end{equation}
where $\cM_{\textrm{ref}}$ is the reference model. When it extends to multilingual, we jointly optimise the DPO loss across multiple languages,

\begin{equation}
\begin{aligned}
    \cL_{\textrm{mpa}}(\cM_\theta)  = \frac{1}{|L|} \sum_{l \in L} \cL_{\textrm{DPO}}(\cM_\theta,\cD^l) .
\end{aligned}
    \label{eq-mpa}
\end{equation}
Compared to mono-lingual alignment, alignment in a multilingual context presents additional challenges, such as negative inter-language interference \citep{Dang2024RLHFCanSpeak, Yu2020Gradientsurgerymulti}. The objective of this paper is to mitigate this problem by designing a data filtering algorithm to filter data that is prone to negative inter-language interference.

\paragraph{Self-rewarding Iterative DPO} Collecting human preferences for training LLMs is a resource-intensive task. Self-rewarding method addresses this challenge by leveraging LLMs to generate  responses and evaluate them to construct their own training preference data
\citep{Yuan2024Selfrewardinglanguage}. 

Specifically, starting from an instruction-tuned model $\cM_{\theta_0}$, we apply an iterative DPO training procedure. At iteration $t > 1$, the model $\cM_{\theta_t}$  is initialised with the previous model $\cM_{\theta_{t-1}}$. For each prompt $x^l$, it generates a set of $k$ candidate responses $\{ y_1^l, \dots, y_k^l \}$. The same model $\cM_{\theta_{t-1}}$ is then used to evaluate each response, resulting in reward scores $\{ r_1^l, \dots, r_k^l \}$. Since the model has the strongest instruction-following ability in English, we use the same English prompt when scoring responses in all languages
 $l \in L$. Preference pairs are then constructed by selecting the highest and lowest-scoring responses, discarding pairs with identical scores.  These pairs are then used to train $\cM_{\theta_t}$. Further details of the prompts are provided in \Cref{apx:prompt}.

%% file: ACL_content/3-method.tex
\section{Conflicting Gradient Filtering for Multilingual Preference Alignment}
\input{ACL_content/fig-congrad}
To address the challenge of conflicting cross-lingual preferences in multilingual alignment, we introduce Conflicting Gradient Filtering (ConGrad), a novel framework designed to select high-quality preference data by minimizing gradient conflicts across languages. Our core idea is to first identify a consensus update direction that is beneficial for all languages and then filter self-generated preference pairs to align with this direction. This ensures a more coherent and effective multilingual optimization trajectory. As illustrated in \Cref{fig:our-method}, ConGrad integrates seamlessly into the self-rewarding loop and comprises two key stages: deriving a consensus gradient and using it for efficient preference data filtering.

\subsection{Deriving a Consensus Gradient}
\label{sec:consensus-gradient}

Our primary objective is to find a single gradient update direction that harmonizes the learning objectives across multiple languages. Such a consensus direction should ideally capture shared preferences between languages while mitigating the negative interference caused by conflicting language-specific gradients. \citet{wang2021gradient} previously observed that higher gradient similarity correlates with improved multilingual performance. Building on this, our goal is to construct a gradient that captures this cross-lingual consensus.

\paragraph{Consensus Gradient Computation}
To achieve this, we first need a stable representation of each language's update direction within a given training iteration $t$. We compute the Exponential Moving Average (EMA) of gradients, $G^l$, for each language $l$ over its minibatches $b^l_{\tau} \sim \cD^l_t$ (see \Cref{alg:pcgrad-filtering}). The EMA gradient is calculated as $G^l_\tau = \gamma G^l_{\tau-1} + (1-\gamma) \grad^l_{\tau}$, where $\grad^l_{\tau}$ is the minibatch gradient and $\gamma$ is a decay factor.

With these stable language-specific gradients, we then need a mechanism to resolve their conflicts. We adapt the principles from Projecting Conflict Gradient (PCGrad)~\citep{Yu2020Gradientsurgerymulti} for this purpose. For each language $l$, we initialize a candidate gradient $\gradpc^{l}$ with its EMA gradient $G^{l}$. We then iteratively project $\gradpc^{l}$ away from the conflicting components of other languages' gradients ($G^{l'}$). A conflict is identified when their cosine similarity is negative. The projection is performed by projecting it onto the normal plane of $G^{l'}$: $\gradpc^{l} \gets \gradpc^{l} - \frac{\gradpc^{l} \cdot G^{l'}}{\|G^{l'}\|^2} \, G^{l'}$.

The final consensus gradient, $\gradpc$, is the sum of these de-conflicted gradients from all languages: $\gradpc = \sum_{l \in L}\gradpc^{l}$. This vector represents a unified update direction that minimizes the interference.
\input{ACL_content/3.1-algo1}

\paragraph{Preference Filtering via Gradient}
In the subsequent training iteration, we use this consensus gradient $\gradpc$ as a reference for quality control. For each newly generated preference sample $i$ in a language $l$, we compute its instantaneous gradient $\grad^l_i$. We then measure its alignment with the consensus direction using cosine similarity, $\cos(\grad^l_i, \gradpc)$. By retaining only the samples with the highest similarity, we curate a dataset that promotes harmonious updates across languages, effectively filtering out those that would introduce conflict.

\input{ACL_content/3.1-algo2}

\subsection{Efficient Gradient EMA via Incremental Low-Rank Updates}
\label{sec:compression}

A practical challenge in our approach is the prohibitive memory cost of storing full EMA gradients for each language, especially for LLMs. To make our method feasible, we maintain a memory-efficient, low-rank approximation of the EMA gradient for each language, stored as a pair of factor matrices $(P^l, Q^l)$.

Our method employs an incremental 'decompress-update-recompress' cycle to update these factors, ensuring the EMA calculation is mathematically sound while managing memory overhead. When a new mini-batch gradient $\grad^l_\tau \in \mathbb{R}^{n\times m}$ arrives for language $l$, the update proceeds in three steps:

\noindent\textbf{Decompress:} First, we reconstruct the full-dimensional EMA gradient from the previous step using its stored factors: 
    $G^l_{\text{ema}, \tau-1} = P^l_{\text{ema}, \tau-1} (Q^l_{\text{ema}, \tau-1})^\top$.
    
\noindent\textbf{Update:} This recovered dense matrix is then updated with the new mini-batch gradient using the standard EMA formula in the full-dimensional space:
    $G'_{\text{updated}} = \gamma G^l_{\text{ema}, \tau-1} + (1-\gamma) \grad^l_\tau$.
    
\noindent\textbf{Re-compress:} To restore memory efficiency, this newly updated, full-dimensional gradient $G'_{\text{updated}}$ is immediately compressed back into a low-rank form. We use power iteration~\citep{bathe1972large} to find its new rank-$r$ factors, $P^l_{\text{ema}, \tau}$ and $Q^l_{\text{ema}, \tau}$, which replace the previous ones in storage. 

Specifically, this process is performed for one matrix at a time. While the gradient for the active matrix is momentarily held in its dense form, the EMA gradients for all other matrices remain compressed, making the peak memory overhead manageable. After training, the final reconstructed gradient $G^l = P^l_{\text{ema}} (Q^l_{\text{ema}})^\top$ is then used for the consensus calculation.

\subsection{Iterative Alignment with ConGrad}
\label{sec:iterative-loop}

We integrate our \textsc{ConGrad} filtering method into an iterative self-rewarding procedure, as detailed in \Cref{alg:modified-training-loop}. The training process begins at iteration $t=1$ by fine-tuning the model on all self-generated preference data. For all subsequent iterations ($t > 1$), the ConGrad module is activated. It first computes the consensus gradient based on the previous iteration's training dynamics and then uses it to filter the newly generated preference dataset, ensuring that only high-quality, low-conflict samples are used for the model update.

Model optimization is performed using DPO. To mitigate the model's tendency towards verbosity, we incorporate a length penalty directly into the loss function \citep{park2024disentangling}. The Length-Penalized DPO (LP-DPO) loss is defined as:
\begin{equation}
  \cL_{\text{LP-DPO}} = -\mathbb{E}_{(x,\ychosen,\yrejected)\sim \cD} \left[ \log \sigma\Big( \beta\, p_m + \alpha\, l_m \Big) \right],
\end{equation}
where $p_m$ is the standard preference margin between the chosen and rejected responses, and $l_m = |\ychosen| - |\yrejected|$ is the length margin. The hyperparameters $\beta$ and $\alpha$ balance the preference-learning objective with the conciseness objective. By minimizing this loss, we guide the model to align with human preferences while maintaining conciseness.

%% file: ACL_content/fig-congrad.tex
\begin{figure}
    \centering
\includegraphics[scale=0.4]{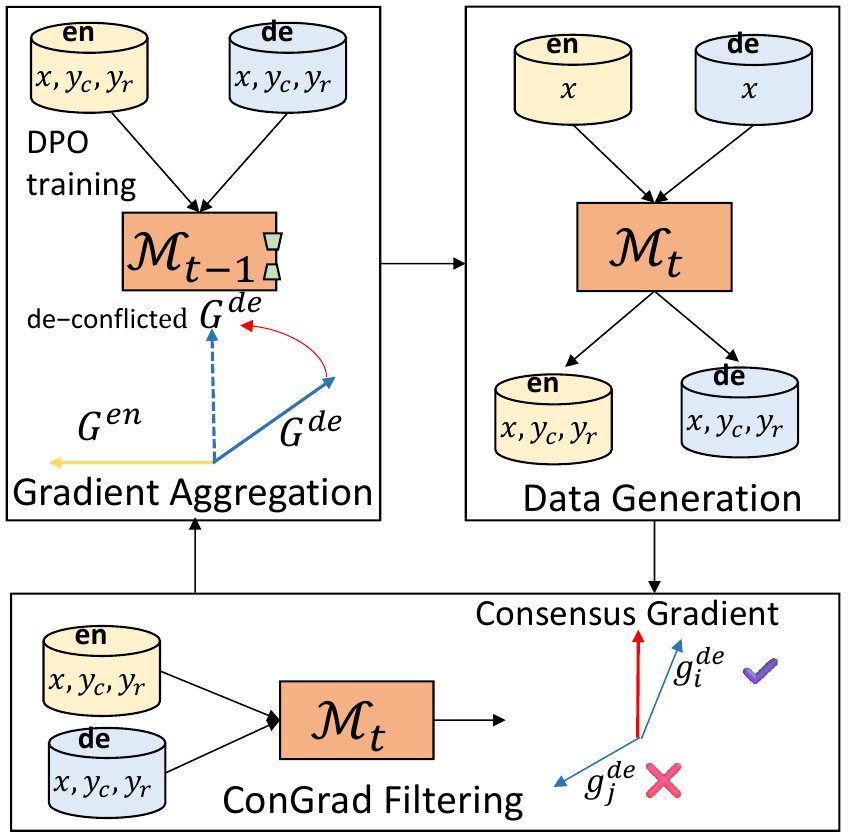}
    \caption{Overview of the multilingual preference alignment framework which consists of three steps: (i) computation of the  cross-lingual consensus gradient via PCGrad with gradient compression for memory efficiency; (ii) generation of synthetic preference data via self-rewarding; and (iii) \textsc{ConGrad} filtering, which selects preference samples based on their gradient similarity to the consensus gradient direction, retaining only those with high similarity ($\grad_i^{de}$ in this example). }
    \label{fig:our-method}
\end{figure}

%% file: ACL_content/3.1-algo1.tex
\begin{figure}

    \begin{minipage}{0.5\textwidth} 
      \begin{algorithm}[H]
        \caption{Conflicting Gradient Filtering}
        \label{alg:pcgrad-filtering}
        \small
        \begin{algorithmic}[1]
          \REQUIRE  Set of languages $L$, gradient EMAs $\{G^l\}_{l \in L}$, self-rewarding dataset $\{\cD^l\}_{l \in L}$ ,  model $\mathcal{M}_{\theta}$

          \STATE Initialise $\gradpc^l \gets G^l \quad \forall l \in L$ 
          \FOR {each language $l \in L$} 
             \FOR {each language $l' \in \textrm{shuffle}(L \setminus \{l\})$}
             \IF {$\gradpc^l . G^{l'} < 0$} 
                \STATE $\gradpc^{l}  \gets \gradpc^{l} - \frac{\gradpc^{l} \cdot G^{l'}}{\|G^{l'}\|^2} \, G^{l'}$ 
              \ENDIF
             \ENDFOR
          \ENDFOR

          \STATE $\gradpc = \sum_{l \in L}\gradpc^{l}$ \hfill // consensus gradient
          \STATE // Filtering step
          \FOR {each language $l \in L$} 
          \FOR{each sample $(x_i^l, \yichosen^l, \yirejected^l) \in \cD^l$}
            \STATE  $\grad_i^l \gets \nabla_{\theta}\cL_\text{LP-DPO}(\cM_{\theta}, (x_i^l, \yichosen^l, \yirejected^l))$
          \ENDFOR
          \STATE {\scriptsize $\mathcal{S}_{\text{selected}}^l \gets \text{ArgTopK}_{(x_i^l, \yichosen^l, \yirejected^l) \in \cD^l} \cos(\grad_i^l, \gradpc)$}
          \ENDFOR
          \RETURN $\{\mathcal{S}_{\text{selected}}^l\}_{l \in L}$
        \end{algorithmic}
      \end{algorithm}

    \end{minipage}
\vspace{-1em}
  \end{figure}

%% file: ACL_content/3.1-algo2.tex
\begin{figure}

    \begin{minipage}{0.5\textwidth} 
    \tiny
{\footnotesize
\begin{algorithm}[H]
\caption{Modified Iterative Self-Rewarding with \textsc{ConGrad} Filtering}
\label{alg:modified-training-loop}
\footnotesize
\begin{algorithmic}[1]
\REQUIRE Set of languages $L$, seed model $\mathcal{M}_{\theta_0}$, prompt dataset $\{\cX^l\}_{l \in L}$, total iterations $T$
\STATE initialise $\tilde{G}^l \gets 0$, for all $l \in L$ 
\FOR{$t = 1$ to $T$}
  \FOR{each language $l \in L$}
    \STATE gen responses $\cY^l$ using $\cX^l$ and $\mathcal{M}_{\theta_{t-1}}$
    \STATE gen rewards $\cR^l$ for $\cY^l$ via self-eval
    \STATE { construct  $\cD^l_{t}$ via pref  pairs in $\cY^l$ and rewards $\cR^l$ }
  \ENDFOR
  \IF{$t > 1$}
    \STATE $\{S_{\text{selected}}^l\}_{l \in L} \gets$ \\ 
    {\scriptsize \textsc{Algorithm 1}$(L,\{\tilde{G}^l * P\}_{l \in L}, \{\cD^l_{t}\}_{l \in L}, \mathcal{M}_{\theta_{t-1}})$}
  \ELSE
    \STATE $\{S_{\text{selected}}^l\}_{l \in L} \gets \{\cD^l_{t}\}_{l \in L}$
  \ENDIF
  \STATE $\mathcal{M}_{\theta_{t}} \gets \mathcal{M}_{\theta_{t-1}}$
  \FOR{ minibatches $ b^l_{\tau} \in \{S_{\text{selected}}^l\}_{l \in L}$ }
     \STATE  $\grad^l_{\tau} \gets\nabla_{\theta_{\tau}} \cL_\text{LP-DPO}(\cM_{\theta}, b^l_{\tau})$
    \STATE $\mathcal{M}_{\theta_{t}} \gets \textrm{UpdateModel}(\mathcal{M}_{\theta_{t}}, \grad^l_{\tau})$
    \STATE $G^l_{\text{old}} \gets P^l_{\text{ema}} (Q^l_{\text{ema}})^\top$ \COMMENT{Decompress}
    \STATE $G'_{\text{updated}} \gets \gamma G^l_{\text{old}} + (1-\gamma) \grad^l_\tau$ \COMMENT{Update}
    \STATE $P^l_{\text{ema}}, Q^l_{\text{ema}} \gets \text{PowerIterationCompress}(G'_{\text{updated}}, r)$ 
 \ENDFOR

\ENDFOR
\RETURN $\mathcal{M}_{\theta_T}$
\end{algorithmic}
\end{algorithm}
}
\end{minipage}
\vspace{-1em}
\end{figure}

%% file: ACL_content/experiment.tex
\section{Experimental Setup}

We evaluate our self-rewarding multilingual LLM framework on several benchmarks and using different gradient filtering strategies. In particular, our experiments are designed to investigate the following three research questions:
\begin{itemize}[noitemsep,topsep=0pt,parsep=0pt,partopsep=0pt,leftmargin=13pt]
  \item \textbf{RQ1}: Is iterative self-rewarding effective for multilingual LLM alignment?
  \item \textbf{RQ2}: Can our \textsc{ConGrad} filtering algorithm for preference data mitigate negative interference in multilingual LLM alignment?
  \item \textbf{RQ3}: How do different preference data filtering methods compare in terms of their effectiveness and characteristics?
\end{itemize}
\vspace{-1mm}
\paragraph{Models} We use widely adopted instruct version of \texttt{llama3-8b} and \texttt{gemma2-2b} as seed models for self-rewarding. Both models support multiple languages, but the capabilities are very uneven between languages. We chose two different sizes of models to see how the various algorithms perform in different parameter sizes.

\noindent\textbf{Implementation Details} In each round, for each prompt, we generate and score four responses and construct preference pairs. For preference data, \textbf{we filter the top 50\% for each language} based on metrics and then perform one epoch of DPO training to ensure data quality and diversity. More details are in Appendix~\ref{apx:Implementation Details}.

\vspace{-1mm}
\subsection{Datasets and Metrics}
\vspace{-1mm}
{\paragraph{Training Dataset} is based on AlpaGasus \citep{chen2024alpagasus}, which contains 9K high-quality English instruction following data filtered from the 52K Alpaca dataset \citep{alpaca}. We randomly sample 1K prompts from AlpaGasus and use Google Translate to translate them into 9 languages: Italian (it), Chinese (zh), Portuguese (pt), Korean (ko), Spanish (es), German (de), Arabic (ar), Japanese (jp), and French (fr). For our multilingual experiments, we split the 1K prompts equally into 10 non-overlapping partitions, i.e., 100 prompts per language. For monolingual experiments, we use the full 1K prompts.}

{\noindent\textbf{Evaluation Datasets} include:
  (1) \textit{aya evaluation suite} for instruction-following \citep{singh-etal-2024-aya}. \textit{aya evaluation suite} contains multilingual open-ended conversation-style prompts to evaluate multilingual open-ended generation quality. It is a high-quality prompt-based benchmark that contains translations and edits by human experts.
  (2) \textit{Global MMLU} and multilingual version of \textit{ARC challenge} for alignment tax \citep{singh2024global}. \textit{Global MMLU} improves upon previous translated MMLU variants by incorporating human-verified translations and annotating subsets for cultural sensitivity. This enables robust evaluation of LLMs across both culturally agnostic and culturally sensitive tasks. \textit{ARC challenge} is a benchmark of grade-school science questions that are specifically selected to be unsolvable by simple retrieval or co-occurrence methods, requiring advanced reasoning and knowledge understanding from models \citep{Clark2018ThinkYH,Lai2023OkapiInstructiontuned}. }

{\noindent\textbf{Metrics} include:
  (1) \textit{Head-to-Head win rate} on \textit{aya evaluation suite}. We utilize GPT-4o to compare the model after self-rewarding alignment during different iterations and the original base model, and calculate the win rate of the model after training over the seed model. 
  (2) \textit{5-shot Accuracy} on \textit{Global MMLU} and multilingual version of \textit{ARC challenge}.}

\input{ACL_content/tab-instruction-following}
\subsection{Baselines} 

We compare our approach, \textsc{ConGrad}, against multiple baseline methods:
\begin{itemize}[noitemsep,topsep=0pt,parsep=0pt,partopsep=0pt,leftmargin=13pt]
\item \textbf{Monolingual Preference Alignment (\textsc{Mono})} We perform self-rewarding using a full 1K prompt dataset for each language. This baseline serves as a reference point to multilingual baselines and provides evidence of negative interference if high-resource languages perform worse in the multilingual setting.
\item \textbf{Multilingual Preference Alignment on Full Data (\textsc{Mult-Full})} We perform self-rewarding across 10 languages, on the full preference dataset containing 100 prompts per language (1K total).
\item \textbf{Multilingual Preference Alignment with Filtering Data} We compare our approach against several filtering techniques for DPO training: (i) Random filtering (\textbf{\textsc{Rand}}), where we report average performance of 3 runs. (ii) Length margin filtering, based on the length gap between chosen and rejected responses. We explore two variants: maximum length margin (\textbf{\textsc{Max-Len}}) and minimum length margin (\textbf{\textsc{Min-Len}}). (iii)  Reward margin filtering based on the self-reward scores for the two responses. It can be viewed as the model's confidence in a sample, where a large margin indicates high confidence. We consider two variants: maximum reward margin (\textbf{\textsc{Max-Reward}}) and min reward margin (\textbf{\textsc{Min-Reward}}). (iv) Our \textbf{\textsc{ConGrad}} method, where we retain the top scoring samples based on the gradient similarity (\textbf{\textsc{Max-ConGrad}}). We also compare with a variant that retains the bottom-scoring samples 
(\textbf{\textsc{Min-ConGrad}}).
\item \textbf{Task-Level Data Scheduling Strategies:}
Uncertainty-Aware Balancing \textbf{(UAB)} uses uncertainty as the metric to adjust the sampler schedule in multilingual tasks. Based on the work of UAB, we use the self-reward preference margin as a proxy for model uncertainty in each language and adjust data sampling probabilities accordingly. Language Curriculum Learning \textbf{(LCL)} found that in multilingual tasks, when the amount of data is unbalanced, it is a good strategy to learn High-resource languages first and then all languages \citet{choi2023order}. Inspired by LCL, we create a curriculum by first training on dominant languages (determined by Global MMLU scores) and then on all languages.

\end{itemize}

\input{ACL_content/tab-llama-tax}

\input{ACL_content/tab-gemma-tax}

\section{Experimental Results}

\vspace{-1mm}
\subsection{Instruction Following Performance}
\vspace{-1mm}

\noindent\textbf{Iterative Self-Rewarding Improves Multilingual Alignment}
As illustrated in Table~\ref{table: main results}, most variants of the self-rewarding procedure yield a sustained increase in win rates over successive iterations, demonstrating that the iterative self-rewarding strategy consistently improves instruction-following performance across different languages.
In the Gemma experiments, our methods exhibited a decline in performance at the fifth round of iteration, suggesting that the benefits of self-rewarding may saturate or reverse due to overfitting.

\noindent\textbf{\textsc{ConGrad} Filtering Enhances Self-Rewarding and Outperforms Other Filtering Methods}
Compared to the vanilla self-rewarding algorithm \textsc{Mult-Full}, \textsc{ConGrad} further improves performance (73.22\% vs 66.52\% on Llama and 69.22\% vs 59.53\% on Gemma). This indicates two points: first, negative cross-lingual interference is present during multilingual alignment; second, our filtering method, which discards samples with low gradient similarity, can effectively mitigate this interference. While our method outperforms other filter methods, we also observe that \textsc{Max-Reward} is a strong baseline, because the base models have modest capacity, leading to noise during scoring. The method reduces noise in preference data by selecting samples with the highest confidence (i.e., \textsc{Max-Reward}). In contrast, filtering based on length differences only yields limited gains. Furthermore, task-level scheduling methods such as UAB and LCL only show limited improvement, which corresponds to a prevailing hypothesis that the quality of individual samples is often more crucial than the sheer quantity of data~\citep{chen2024alpagasus, zhou2023lima}.

\noindent\textbf{Coexistence of Positive Transfer and Negative Interference Across Languages}
Beyond negative interference, the results also indicate positive cross-lingual transfer. For instance, in the Gemma results, compared to the \textsc{Mono}, \textsc{Mult-Full} significantly boosts performance for several languages. The average win rates on the best epoch increase from 49.14\% to 59.53\%, even though the data volume is controlled to be the same between the multilingual and monolingual settings. In more detailed results in \Cref{apx:addresult}, we find that these improvements come mainly from underperforming languages such as Arabic, Korean, and German. This suggests that alignment improvements can largely stem from positive transfer from better-performing languages for languages with initially low performance.


\input{ACL_content/fig-percentage.tex}

\input{ACL_content/tab-unseen}

\input{ACL_content/table_low_resource}

\noindent\textbf{Impact of Gradient Filter Percentage}
The percentage of the filter serves as an important hyperparameter that controls the strength of the filter algorithm. Ideally, the algorithm should not be too sensitive to the filtering strength within a certain range. In Figure~\ref{fig:percentage}, we report the win rate resulting from retaining different percentages of preference data while performing filtering. As shown, for Llama, retaining the top 50\% of the data for each language yields the best performance but is overall insensitive to the filter strength. For Gemma, on the other hand, retaining 25\% yields the best results at the best iteration round. This may be because the performance in Gemma is weaker compared to Llama, so there is more noise in the process of reward computation, and increasing the filter strength improves the performance to some extent.
\vspace{-1mm}

\subsection{Alignment Tax}
\vspace{-1mm}

Previous work has shown that after alignment training, such as DPO or RLHF, the model may forget some of its knowledge, which in turn leads to some degradation of the basic capabilities, also known as the alignment tax. Ideally, alignment algorithms should not produce much alignment tax.
To test whether our algorithm suffers from alignment tax, we use two datasets for testing, \textit{Global MMLU} and \textit{ARC challenge}. For \textit{Global MMLU}, we test the same ten languages. For \textit{ARC challenge}, we test the other eight languages due to their lack of Korean and Japanese data. From the experimental results of Table~\ref{table: tax llama} and Table~\ref{table: tax gemma}, we can find that our algorithm even slightly improves the model's performance on both datasets without incurring an alignment tax. This may stem from the fact that the increase in the model's instruction following capability boosts its 5-shot in-context learning performance.

%% file: ACL_content/tab-instruction-following.tex
\begin{table*}[t]
    \centering
    \resizebox{0.9\textwidth}{!}{ 
    \begin{tabular}{l|rrrrr|rrrrr}
    \toprule

        \multirow{2}{*}{\textbf{Method/Round}} &  \multicolumn{5}{c|}{\textbf{Llama}}  & \multicolumn{5}{c}{\textbf{Gemma}}  \\ 
        \cmidrule(r){2-6} \cmidrule(r){7-11}
         & \multicolumn{1}{c}{\textbf{1}} & \multicolumn{1}{c}{\textbf{2}} & \multicolumn{1}{c}{\textbf{3}} & \multicolumn{1}{c}{\textbf{4}} & \multicolumn{1}{c|}{\textbf{5}} & \multicolumn{1}{c}{\textbf{1}} & \multicolumn{1}{c}{\textbf{2}} & \multicolumn{1}{c}{\textbf{3}} & \multicolumn{1}{c}{\textbf{4}} & \multicolumn{1}{c}{\textbf{5}} \\
         \midrule
        \textsc{Mono} & 56.25 & 61.16 & 63.80 & 65.79 & 66.08 & 49.14 & 46.98 & 43.01 & 39.09 & 38.11    \\ 
        \midrule
        \textsc{Mult-Full} & 57.57 & 60.94 & 61.75 & 64.42 & 66.52 & 57.80 & 57.93 & 57.31 & 57.51 & 59.53 \\ 
        \midrule
        \textsc{UAB} & 57.57 & 61.51 & 63.04 & 64.51 & 66.54 & 57.80 & 58.73 & 59.63 & 60.61 & 60.85 \\  
        \textsc{LCL} & 57.57 & 60.48 & 63.96 & 65.76 & 66.67 & 57.80 & 59.65 & 60.04 & 60.52 & 62.51 \\
        \midrule
        \textsc{Rand} & 57.57 & 61.56 & 64.06 & 65.21 & 66.32 & 57.80 & 56.27 & 57.98 & 54.25 & 57.25 \\  
        \textsc{Min-Len} & 57.57 & 58.19 & 59.13 & 61.55 & 63.57 & 57.80 & 51.56 & 58.94 & 52.66 & 54.29  \\
        \textsc{Max-Len} & 57.57 & 60.33 & 59.88 & 61.11 & 67.03 & 57.80 & 58.83 & 58.84 & 58.94 & 59.52 \\ 
        \textsc{Min-Reward} & 57.57 & 59.40 & 57.11 & 58.32 & 57.27 & 57.80 & 50.09 & 45.18 & 44.67 & 40.90 \\ 
        \textsc{Max-Reward}& 57.57 & \textbf{64.66} & \underline{67.23} & \underline{70.39} & \underline{70.64} & 57.80 & \textbf{61.49} & \underline{63.73} & \underline{67.72} & \underline{61.82} \\
        \midrule
        \textsc{Min-ConGrad} & 57.57 & 56.14 &  58.03 & 57.94 & 64.25 & 57.80 & 51.69 & 56.69 & 54.35 & 56.34 \\ 
        \textsc{Max-ConGrad} & 57.57 & \underline{64.93} & \textbf{66.58} & \textbf{69.99} & \textbf{73.22} & 57.80 & \underline{60.31} & \textbf{65.54} & \textbf{69.22} & \textbf{66.36} \\
        \bottomrule
    \end{tabular}

}
\caption{\label{table: main results} The average win rates of self-rewarding variants on aya evaluation suite. \textbf{Bold} indicates the best, \underline{underline} the second-best.}

\end{table*}
\vspace{-1mm}

%% file: ACL_content/tab-llama-tax.tex
\begin{table*}[t]
    \centering
    \resizebox{0.8\textwidth}{!}{ 
    \begin{tabular}{lccccc|ccccc}
    \toprule

        \multirow{2}{*}{\textbf{Method/Round}} &\multicolumn{5}{c|}{\textbf{Global MMLU (Seed Model: 48.67)}}  & \multicolumn{5}{c}{\textbf{ARC Challenge (Seed Model: 72.23)}} \\ 
        \cmidrule(r){2-6} \cmidrule(r){7-11}
         & \textbf{1} & \textbf{2} & \textbf{3} & \textbf{4} & \textbf{5} & \textbf{1} & \textbf{2} & \textbf{3} & \textbf{4} & \textbf{5} \\ 
        \midrule
        \textsc{Mult-Full} & 48.85 & 49.01 & 49.14 & 49.27 & 49.47 & 72.41 & 72.45 & 72.51 & 72.45 & 72.57  \\ 
        \midrule
        \textsc{Rand} & 48.85 & 49.13 & 49.16 & 49.16 & 49.23 & 72.41 & 72.40 & 72.52 & 72.36 & 72.42  \\ 
        \textsc{Min-Len} & 48.85 & \textbf{50.40} & \textbf{50.47} & \textbf{50.52} & \textbf{50.62} & 72.41 & \textbf{72.82} & \underline{72.84} & \textbf{72.99} & \textbf{72.99}  \\ 
        \textsc{Max-Len} & 48.85 & \underline{50.25} & \underline{50.17} & \underline{49.92} & \underline{49.78} & 72.41 & \underline{72.78} & \textbf{72.85} & \underline{72.81} & \underline{72.79}  \\
        \textsc{Min-Reward} & 48.85 & 49.03 & 49.19 & 49.40 & 49.40 & 72.41 & 72.40 & 72.54 & 72.60 & 72.66  \\ 
        \textsc{Max-Reward} & 48.85 & 48.72 & \textcolor{IndianRed3}{48.63} & \textcolor{IndianRed3}{48.57} & \textcolor{IndianRed3}{48.28} & 72.41 & 72.29 & \textcolor{IndianRed3}{72.19} & 72.23 & \textcolor{IndianRed3}{72.09}  \\
        \midrule
        \textsc{Min-ConGrad} & 48.85 & 49.00 & 49.07 & 49.15 & 49.11 & 72.41 & 72.40 & 72.49 & 72.32 & 72.41  \\ 
        \textsc{Max-ConGrad} & 48.85 & 49.09 & 49.19 & 49.11 & 48.90 & 72.41 & 72.37 & 72.25 & 72.31 & 72.28 \\
    \bottomrule
    \end{tabular}
}
\caption{\label{table: tax llama} The average accuracy of self-rewarding variations built on Llama. \textcolor{IndianRed3}{Red numbers} indicate lower performance than the seed model.  \textbf{Bold} indicates the best, \underline{underline} the second-best.}
\vspace{-1em}
\end{table*}

%% file: ACL_content/tab-gemma-tax.tex
\begin{table*}[t]
    \centering
    \resizebox{0.8\textwidth}{!}{ 
    \begin{tabular}{lccccc|ccccc}
    \toprule

        \multirow{2}{*}{\textbf{Method/Round}} &\multicolumn{5}{c|}{\textbf{Global MMLU (Seed Model: 38.88)}}  & \multicolumn{5}{c}{\textbf{ARC Challenge (Seed Model: 62.85)}} \\ 
        \cmidrule(r){2-6} \cmidrule(r){7-11}
         & \textbf{1} & \textbf{2} & \textbf{3} & \textbf{4} & \textbf{5} & \textbf{1} & \textbf{2} & \textbf{3} & \textbf{4} & \textbf{5} \\ \midrule
        \textsc{Mult-Full} & 40.03 & 39.78 & 39.50 & 40.07 & 40.48 & 63.65 & 63.52 & 63.81 & 63.81 & 64.03  \\ \midrule
        \textsc{Rand} & 40.03 & 39.50 & 39.29 & 39.05 & 39.73 & 63.65 & 63.42 & 63.03 & \textcolor{IndianRed3}{62.68} & 63.12  \\ 
        \textsc{Min-Len} & 40.03 & 39.39 & 39.31 & \textcolor{IndianRed3}{37.41} & \textcolor{IndianRed3}{37.00} & 63.65 & 63.08 & 63.10 & \textcolor{IndianRed3}{61.93} & \textcolor{IndianRed3}{61.28}  \\ 
        \textsc{Max-Len} & 40.03 & \underline{40.83} & \underline{41.35} & \underline{41.80} & \underline{41.80} & 63.65 & \underline{63.94} & \underline{64.17} & \underline{64.22} & \underline{64.35}  \\ 
        \textsc{Min-Reward} & 40.03 & 39.78 & \textcolor{IndianRed3}{38.68} & \textcolor{IndianRed3}{38.32} & \textcolor{IndianRed3}{36.40} & 63.65 & 63.41 & 63.04 & 62.91 & \textcolor{IndianRed3}{61.99}  \\ 
        \textsc{Max-Reward} & 40.03 & \textbf{40.87} & \textbf{41.93} & \textbf{41.97} & \textbf{42.48} & 63.65 & \textbf{63.96} & \textbf{64.35} & \textbf{64.45} & \textbf{64.48}  \\ \midrule
        \textsc{Min-ConGrad} & 40.03 & 40.29 & 39.54 & 40.98 & \textcolor{IndianRed3}{38.81} & 63.65 & 63.56 & 63.29 & 63.61 & \textcolor{IndianRed3}{62.81}  \\ 
        \textsc{Max-ConGrad} & 40.03 & 39.75 & 39.93 & 39.77 & 39.18 & 63.65 & 63.58 & 63.58 & 63.42 & 63.38 \\ \bottomrule
    \end{tabular}
}
\caption{\label{table: tax gemma} The average accuracy of self-rewarding variations built on Gemma. \textcolor{IndianRed3}{Red numbers} indicate lower performance than the seed model.  \textbf{Bold} indicates the best, \underline{underline} the second-best.}

\end{table*}

%% file: ACL_content/fig-percentage.tex
\begin{figure}[t] 
    \vspace{-1em}
    \centering
    \includegraphics[scale=0.3]{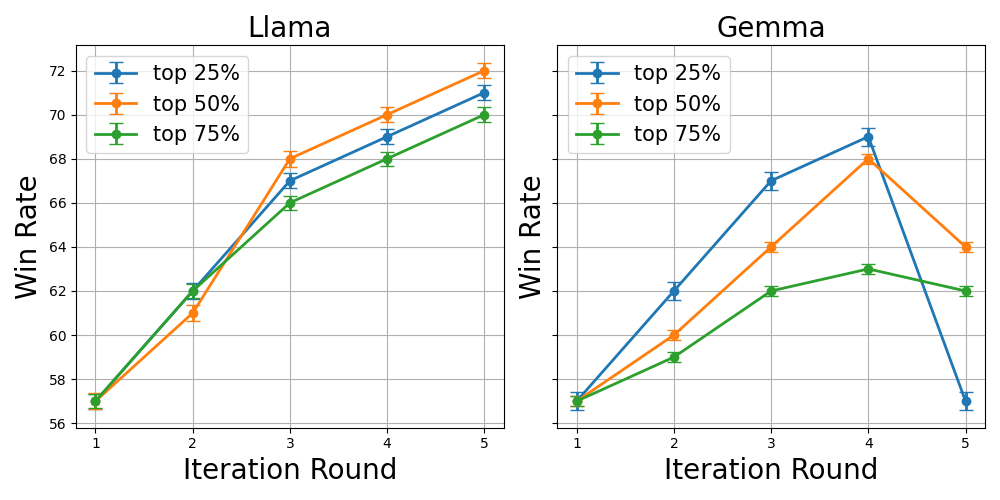}
    \vspace{-2em}
    \caption{Win rate of different filter percentages.}
    \label{fig:percentage}
    \vspace{-1em}
\end{figure}

%% file: ACL_content/tab-unseen.tex
\begin{table*}[t]
    \centering
    \resizebox{\textwidth}{!}{ 
    \begin{tabular}{l|ccccc|ccccc|ccccc}
    \toprule
        \multirow{2}{*}{\textbf{Method/Round}} & \multicolumn{5}{c}{\textbf{Hindi}} &\multicolumn{5}{c}{\textbf{Vietnamese}}  & \multicolumn{5}{c}{\textbf{Russian}}  \\ 
        \cmidrule(r){2-6} \cmidrule(r){7-11} \cmidrule(r){12-16}
         & \textbf{1} & \textbf{2} & \textbf{3} & \textbf{4} & \textbf{5} & \textbf{1} & \textbf{2} & \textbf{3} & \textbf{4} & \textbf{5} & \textbf{1} & \textbf{2} & \textbf{3} & \textbf{4} & \textbf{5}\\ \midrule
         \multicolumn{16}{c}{\textbf{\texttt{Llama3-8B}}}  \\

        \textsc{Mult-Full}  & 54.50 & 57.50 & 61.25 & 62.25 & 65 & 59.25 & 62.00 & 65.00 & 65.00 & 69.25 & 55.53 & 58.50 & \underline{63.00} & 64.75 & 66.25 \\ 
        \textsc{Max-Reward}  & 54.50 & \textbf{66.00} & \underline{67.25} & \underline{69.25} & \underline{71.00} & 59.25 & \underline{63.00} & \underline{68.50} & \underline{70.75} & \textbf{73.50} & 55.53 & \textbf{61.70} & \textbf{68.00} & \textbf{70.25} & \underline{71.25}  \\ 
        \textsc{Max-ConGrad}  & 54.50 & \underline{63.75} & \textbf{68.00} & \textbf{70.25} & \textbf{73.75} & 59.25 & \textbf{63.75} & \textbf{71.75} & \textbf{71.25} & \underline{72.50} & 55.53 & \underline{61.00} & \underline{67.50} & \underline{69.75} & \textbf{71.50}  \\
        \midrule
        \midrule
        \multicolumn{16}{c}{\textbf{\texttt{Gemma2-2B}}} \\
        \textsc{Mult-Full}   & 54.52 & \textbf{59.55} & 55.78 & 55.78 & 56.03 & 56.50 & 54.00 & 55.75 & 55.00 & 58.25 & 59.05 & 61.81 & 58.29 & 63.32 & 61.81 \\ 
        \textsc{Max-Reward}  & 54.52 & 54.52 & \underline{61.56} & \underline{63.07} & \underline{57.54} & 56.50 & \underline{60.00} & \underline{62.50} & \underline{63.00} & \textbf{66.50} & 59.05 & \underline{63.32} & \textbf{69.85} & \underline{65.83} & \underline{64.32}  \\ 
        \textsc{Max-ConGrad}  & 54.52 & \underline{55.03} & \textbf{62.81} & \textbf{63.82} & \textbf{58.04} & 56.50 & \textbf{62.00} & \textbf{64.00} & \textbf{71.75} & \underline{65.50} & 59.05 & \textbf{66.08} & \underline{65.33} & \textbf{69.60} & \textbf{67.09}  \\ 
        \bottomrule
    \end{tabular}
}

\caption{\label{table:llama unseen} The average win rates of self-rewarding variants on unseen languages from aya evaluation suite. \textbf{Bold} indicates the best, \underline{underline} the second-best.}
\vspace{-9mm}
\end{table*}

%% file: ACL_content/table_low_resource.tex
\begin{table}[t]
\vspace{5mm}
\resizebox{0.5\textwidth}{!}{ 
    \centering

    \label{tab:underrepresented}
    \begin{tabular}{lccccc}
    \toprule
    \textbf{Method} & \textbf{avg of others} & \textbf{lv} & \textbf{lt} & \textbf{sl} & \textbf{sk} \\
    \midrule
    \textsc{Mult-Full}   & 67.7 & 69.5  & 74.25 & 74.25 & 68.0  \\
    \textsc{Reward}  & 69.5 & 74.25 & 73.0  & 74.25 & 75.75 \\
    \textsc{ConGrad} & \textbf{73.2} & \textbf{76.5}  & \textbf{75.5}  & \textbf{81.0}  & \textbf{78.75} \\
    \bottomrule
    \end{tabular}
}
\caption{\label{table:low resource}Win rates on underrepresented languages.}
\vspace{-5mm}
\end{table}

%% file: ACL_content/5-analysis.tex
\vspace{-2mm}
\subsection{Analysis}

\vspace{-1mm}
\paragraph{Performance on Unseen Languages}  In some extreme cases, we may need to use certain languages that are not included in the self-rewarding process. To verify the usability of our model in such cases, we randomly selected three other widely varying languages, Hindi, Vietnamese and Russian, for validation. According to Table~\ref{table:llama unseen}, multilingual self-rewarding generalizes well to unseen languages, indicating a positive transfer between languages. In addition, better results can still be achieved using \textsc{ConGrad}, illustrating the generalizability of our approach.
\vspace{-1mm}
\paragraph{Gains in Underrepresented Languages}
Current LLMs often lack capability in many underrepresented languages.
To test the robustness of \textsc{ConGrad} in these settings, we added four new languages with minimal \texttt{llama3-8b} tokenizer support (around 800+ tokens), Latvian (lv), Lithuanian (lt), Slovenian (sl), and Slovak (sk), to the original ten languages for alignment. The results, summarized in \Cref{table:low resource}, show that \textsc{ConGrad} improves alignment in these languages while maintaining high performance of others. \textsc{ConGrad} consistently outperformed the baselines in all four underrepresented languages and on the original ten. These findings suggest CONGRAD enhances alignment even with limited base model support, indicating its generalization potential.

%% file: ACL_content/7-conclusion.tex
\section{Conclusion}
In this work, we propose \textsc{ConGrad}, a multilingual LLM alignment framework with conflicting gradient filtering, which avoids the reliance on external annotations. Our approach iteratively generates and scores preference data and introduces a PCGrad-based gradient filtering strategy to mitigate negative cross-lingual interference. Extensive experiments on LLaMA3 and Gemma2 show that our approach significantly improves instruction following for underrepresented languages, maintains performance for mainstream languages, and generalizes to unseen languages. Furthermore, our analysis highlights the dual role of positive transfer and negative interference in multilingual training, suggesting that careful data selection is essential to fully unlock the multilingual potential of aligned LLMs.

%% file: ACL_content/limitation.tex
\section{Limitations}
While \textsc{ConGrad} demonstrates strong empirical performance in mitigating negative cross-lingual interference, several limitations remain. First, although sublinear gradient compression significantly reduces memory overhead, it introduces approximation noise that may impact the accuracy of gradient filtering. Second, our experiments are limited to typologically diverse but relatively balanced languages; generalization to highly imbalanced or severely low-resource language scenarios remains an open challenge. However, this may be more of a problem with the pre-training stage than with the alignment stage since the LLM's most basic language capabilities still require the support of pre-trained data. Lastly, extending the framework to support dynamic filtering during training, rather than round-level static selection, could enable finer-grained control over training dynamics but might also introduce additional costs.

Future work can explore hybrid filtering strategies that combine gradient-based and reward-based signals to improve sample selection. Additionally, integrating cross-lingual alignment objectives directly into the optimization process, beyond data filtering, may further enhance multilingual alignment. 

%% file: ACL_content/Appendix.tex
\input{ACL_content/6-relatedwork}

\section{Implementation Details}
\label{apx:Implementation Details}

To optimize model performance, we conducted systematic hyperparameter tuning during training. We tuned key hyperparameters, including the learning rate, compression dimension, and the length penalty strength $\alpha$ used in DPO training. Specifically, we searched the learning rate over [$2 \times10^{-6}$, $1 \times10^{-6}$, $5 \times10^{-7}$], while the batch size was fixed at 16 and the compression rank was fixed at 64 to balance GPU memory constraints and training stability, which is similar with LoRA~\citep{hulora}. When performing power iteration, we set the number of iterations to 3 to ensure a balance between precision and efficiency. The length penalty strength parameter was explored within [$0.02$, $0.01$, $0.005$].

For LLama, all training was performed on 4 A100 GPUs, and for Gemma, all training was performed on 2 A100 GPUs. We use the openrlhf framework~\citep{hu2024openrlhf} for DPO training and vLLM~\citep{kwon2023efficient} for inference.
\section{Further Analysis}
\subsection{Gradient Approximation Quality and Sensitivity Analysis}

\textbf{Approximation Quality} We first evaluated the approximation error of our gradient compression technique. For the chosen rank of $r=64$, the recovered low-rank gradients exhibited a strong directional correlation with the original full-rank gradients, achieving an average cosine similarity of approximately 0.8. We determined that this level of directional approximation was adequate to guide the data filtering process effectively without requiring the full, computationally expensive gradients.

\textbf{Sensitivity to Compression Ratio} To investigate the model's sensitivity to the degree of compression, we conducted an ablation study on the \texttt{Llama3-8B} model with multiple varied ranks. The results, summarized in Table~\ref{tab:sensitivity_analysis}, demonstrate that the performance of \textsc{ConGrad} is robust to the choice of rank within a reasonable range ($r \ge 8$), showing only marginal fluctuations. A significant performance degradation was observed only when the compression became highly aggressive ($r=4$). We attribute this robustness to our data selection mechanism, which filters the top 50\% of data based on gradient scores. This approach is likely resilient to minor inaccuracies in gradient approximation, as the relative ranking of the most beneficial samples is largely preserved even with compressed gradients.

\begin{table}[htbp]
\centering
\caption{Performance sensitivity of CONGRAD on Llama3-8B with varying compression ranks ($r$).}
\label{tab:sensitivity_analysis}
\begin{tabular}{lc}
\toprule
\textbf{Rank ($r$)} & \textbf{Average Win Rate (\%)} \\
\midrule
64 & 73.2 \\
32 & 72.6 \\
16 & 72.3 \\
8  & 71.9 \\
4  & 67.5 \\
\bottomrule
\end{tabular}
\end{table}
\input{ACL_content/apx-prompts}

\section{Addition Results}
\label{apx:addresult}
\input{ACL_content/tab-llama-lang-detail}

\input{ACL_content/tab-gemma-lang-detail}

%% file: ACL_content/6-relatedwork.tex
\section{Related Work}
\vspace{-1.5mm}
\paragraph{Multilingual Preference Optimization} 
Recent work has explored bridging the performance gap between high- and low-resource languages through multilingual preference optimization. \citet{Li2024ImprovingcontextLearning, Zhang2024EnhancingMultilingualCapabilities} align internal representations and outputs using contrastive learning and instruction tuning. \citet{Yang2025LanguageImbalanceDriven} leverage language imbalance as a natural preference signal for iterative self-improvement across languages, while \citet{Yang2025ImplicitCrossLingual} transfer implicit rewards from English-aligned models to other languages without explicit multilingual preference data. \citet{Dang2024RLHFCanSpeak} show that cross-lingual transfer emerges from multilingual preference training, with online methods outperforming offline ones. \citet{Zhang2024EnhancingMultilingualCapabilities} propose a self-distillation method to improve multilingual generation by utilizing strong responses in resource-rich languages. Meanwhile, \citet{Gureja2024MRewardBenchEvaluating} reveal significant performance gaps in multilingual reward models, emphasizing the need for more robust evaluation and alignment across languages.

\vspace{-1.5mm}
\paragraph{Synthetic Multilingual Dataset Creation} Many studies create multilingual preference data via translation from English instructions \citep{Lai2023OkapiInstructiontuned, shaham2024multilingual, mondshine2025beyond}. Okapi \citep{Lai2023OkapiInstructiontuned} translates English prompts into 26 languages and ranks responses using GPT-3.5. \citet{shaham2024multilingual} find that adding just a few multilingual examples improves cross-lingual generalization. \citet{mondshine2025beyond} propose selective pre-translation of prompt components, improving performance across 35 languages. Other work leverages language imbalance as a heuristic to generate multilingual preference pairs \citep{Yang2025LanguageImbalanceDriven}.

\vspace{-1.5mm}
\paragraph{Instruction and Preference Data Filtering}
Instruction data filtering methods aim to identify the most useful samples for fine-tuning. LESS \citep{Xia2024LESSSelectingInfluential} selects influential instructions by estimating gradient similarity with few-shot targets, while DART-Math \citep{tong2024dart} prioritizes difficult queries during data synthesis to enhance reasoning. Alpagasus \citep{chenalpagasus} filters low-quality instruction-response pairs using GPT-based scoring. \citet{Wu2024bestbothworlds} balance quality and diversity via a graph-based selector. For preference data, fDPO \citep{morimura2024filtered} introduces reward-model-based filtering during DPO training, showing that preference data quality critically impacts alignment. Overall, most prior work focuses on instruction data, while systematic filtering of preference data remains underexplored.

%% file: ACL_content/apx-prompts.tex
\section{Self-rewarding Iterative DPO Prompts}
\label{apx:prompt}
\begin{tcolorbox}[colback=gray!5, colframe=gray!60, title=Response Evaluation Prompt for Self-Rewarding, 
  fontupper=\ttfamily, 
  enhanced jigsaw, 
  breakable, 
  sharp corners,
  boxrule=0.5pt]
Review the user's question and the corresponding response using the **additive 5-point scoring system** described below. Points are accumulated based on the satisfaction of each criterion: 

- Add 1 point (total Score: 1) if the response is relevant and provides some information related to the user's inquiry, even if it is incomplete or contains some irrelevant content. 

- Add another point (total Score: 2) if the response addresses a substantial portion of the user's question, but does not completely resolve the query or provide a direct answer. 

- Add another point (total Score: 3) if the response answers the basic elements of the user's question in a useful way, regardless of whether it seems to have been written by an AI Assistant or if it has elements typically found in blogs or search results. 

- Add another point (total Score: 4) if the response is clearly written from an AI Assistant's perspective,  addressing the user's question directly and comprehensively, and is well-organized and helpful, even if there is slight room for improvement in clarity, conciseness or focus. 

- Add another point (total Score: 5) for a response that is impeccably tailored to the user's question  by an AI Assistant, without extraneous information, reflecting expert knowledge, and  demonstrating a high-quality, engaging, and insightful answer. 

User: {0} 

<response>{1}</response> 

Remember to assess from the AI Assistant perspective. To evaluate the response in alignment with this additive scoring model, we'll systematically attribute points based on the outlined criteria.

After examining the user's instruction and the response: 

- Briefly analyse the response in *English*, *up to 100 words*, which is a *strict limit*, from the AI Assistant perspective.

- ** Conclude with the score of the response *strictly using English* and the format: “Score: <total points>” **

- The score should be an **integer from 0 to 5**
\end{tcolorbox}

%% file: ACL_content/tab-llama-lang-detail.tex
\begin{table*}[!ht]
    \centering
    \resizebox{\textwidth}{!}{ 
    \begin{tabular}{lcccccccccc|c}
    \toprule
        & it & zh & pt & en & ko  & es & de & ar & ja & fr & avg  \\ \midrule
        \textsc{Mono} & 52.50 & 71.20 & 59.20 & 55.75 & 76.00 & 60.64 & 64.00 & 71.26 & 76.50 & 73.76 & 66.08  \\
        \midrule
        \textsc{Mult-Full} & 61.75 & 64.60 & 66.00 & 54.56 & 75.00 & 67.82 & 63.00 & 63.98 & 74.00 & 74.50 & 66.52  \\ 
        \midrule
        {UAB} & 62.50 & 71.29 & 64.40 & 53.00 & 72.75 & 70.00 & 60.75 & 59.00 & 74.75 & 77.00 & 66.54  \\ 
        {LCL} & 67.00 & 72.09 & 63.40 & 54.00 & 77.25 & 67.75 & 62.50 & 59.00 & 76.25 & 77.50 & 67.67 \\ 
        \midrule
        \textsc{Rand}& 63.50 & 70.88 & 63.20 & 53.37 & 74.75 & 66.83 & 62.25 & 59.72 & 72.75 & 75.99 & 66.32  \\ 
        \textsc{Min-Len} & 58.00 & 69.00 & 65.40 & 53.17 & 70.50 & 58.17 & 58.75 & 64.17 & 71.00 & 67.57 & 63.57 \\        
        \textsc{Max-Len} & 63.25 & 65.86 & 64.20 & 53.17 & 75.50 & 66.83 & 63.00 & 69.49 & 77.00 & 72.03 & 67.03  \\ 
        \textsc{Min-Reward} & 55.75 & 59.44 & 56.20 & 45.63 & 60.00 & 58.25 & 58.75 & 55.91 & 59.25 & 63.61 & 57.27  \\ 
        \textsc{Max-Reward} & \underline{69.75} & \textbf{76.91} & \underline{70.40} & \textbf{58.93} & \underline{78.25} & \underline{71.50} & \underline{66.25} & \underline{59.45} & \underline{76.50} & \underline{78.47} & \underline{70.64}  \\
        \midrule
        \textsc{Min-ConGrad} & 69.00 & 65.26 & 62.80 & 60.12 & 65.00 & 61.31 & 63.50 & 61.00 & 67.00 & 67.50 & 64.25  \\        
        \textsc{Max-ConGrad} & \textbf{71.00} & \underline{74.50} & \textbf{75.40} & \underline{57.14} & \textbf{82.25} & \textbf{72.75} & \textbf{71.50} & \textbf{65.16} & \textbf{82.00} & \textbf{80.50} & \textbf{73.22}  \\
    \bottomrule
    \end{tabular}
}
\caption{\label{table:llama-lang}The win rate of self-rewarding variations of LLama3-8B evaluated on aya evaluation suite for each language in the final round. The best results are highlighted in \textbf{bold}, and the second-best results are highlighted in \underline{underline.}}
\end{table*}

%% file: ACL_content/tab-gemma-lang-detail.tex
\begin{table*}[!ht]
    \centering
    \resizebox{\textwidth}{!}{ 
    \begin{tabular}{lcccccccccc|c}
    \toprule
         & it & zh & pt & en & ko  & es & de & ar & ja & fr & avg  \\ \midrule
        \textsc{Mono} & 43.50 & 62.65 & 49.60 & 55.40 & 45.25 & 50.25 & 46.50 & 31.22 & 46.75 & 60.30 & 49.14  \\ 
        \midrule
        \textsc{mult-full} & 61.00 & 62.45 & 59.00 & 58.40 & 70.00 & 51.25 & 56.75 & 56.85 & 60.50 & 59.05 & 59.53  \\ 
        \midrule
        {UAB} & 67.50 & 59.84 & 58.40 & 59.60 & 70.00 & 51.75 & 56.50 & 59.40 & 65.00 & 60.50 & 60.85  \\ 
        {LCL} & 68.50 & 62.24 & 60.80 & 57.60 & 74.00 & 55.25 & 59.00 & 61.20 & 65.75 & 60.75 & 62.51 \\          
        \midrule 
        \textsc{Rand} & 58.25 & 63.45 & 58.20 & 55.40 & 63.00 & 54.27 & 56.25 & 60.00 & 54.25 & 56.75 & 57.98  \\ 
        \textsc{Min-Len} & 51.01 & 57.29 & 53.20 & 52.78 & 61.75 & 42.50 & 45.00 & 58.10 & 51.75 & 42.25 & 51.56 \\         
        \textsc{Max-Len} & 61.75 & 63.45 & 58.40 & 58.73 & 63.25 & 51.76 & 59.50 & 60.84 & 59.25 & 58.25 & 59.52  \\ 
        \textsc{Min-Reward} & 50.00 & 50.00 & 46.80 & 47.40 & 55.00 & 44.72 & 52.75 & 59.64 & 47.50 & 49.75 & 50.09  \\ 
        \textsc{Max-Reward} & \underline{70.75} & \underline{69.68} & \underline{63.80} & \underline{61.51} & \underline{72.25} & \textbf{69.60} & \underline{64.75} & \textbf{67.07} & \textbf{69.25} & \underline{68.59} & \underline{67.72}  \\    
        \midrule        
        \textsc{Min-ConGrad} & 63.25 & 57.03 & 54.00 & 52.38 & 63.00 & 49.75 & 56.00 & 56.20 & 52.75 & 59.05 & 56.34  \\ 
        \textsc{Max-ConGrad} & \textbf{68.00} & \textbf{72.29} & \textbf{66.40} &\textbf{63.49} & \textbf{76.25} & \underline{69.50} & \textbf{66.25} & \underline{65.35} & \underline{76.00} & \textbf{68.75} & \textbf{69.22}  \\
        \bottomrule
    \end{tabular}
}
\caption{\label{table:llama-lang}The win rate of self-rewarding variations of Gemma2-2B evaluated on aya evaluation suite for each language in the fourth round. The best results are highlighted in \textbf{bold}, and the second-best results are highlighted in \underline{underline.}}
\end{table*}